\documentclass[10pt,twocolumn,letterpaper]{article}

\usepackage{cvpr}              
\usepackage{makecell}
\usepackage[dvipsnames]{xcolor}

\usepackage{amsmath,amsfonts,bm}

\def\eqref#1{equation~\ref{#1}}

\def\1{\bm{1}}

\DeclareMathAlphabet{\mathsfit}{\encodingdefault}{\sfdefault}{m}{sl}
\SetMathAlphabet{\mathsfit}{bold}{\encodingdefault}{\sfdefault}{bx}{n}

\definecolor{cvprblue}{rgb}{0.21,0.49,0.74}
\usepackage[pagebackref,breaklinks,colorlinks,citecolor=cvprblue]{hyperref}

\def\confName{CVPR}
\def\confYear{2024}

\title{Directional Texture Editing for 3D Models}

\author{
    Shengqi Liu\textsuperscript{\rm 1}\thanks{Equal contribution.} \quad
    Zhuo Chen\textsuperscript{\rm 1*} \quad
    Jingnan Gao\textsuperscript{\rm 1} \quad
    Yichao Yan\textsuperscript{\rm 1}\thanks{Corresponding authors} \\
    Wenhan Zhu\textsuperscript{\rm 1} \quad
    Jiangjing Lyu\textsuperscript{\rm 2} \quad
    Xiaokang Yang\textsuperscript{\rm 1} \\
    \textsuperscript{\rm 1} Shanghai Jiao Tong University \quad
    \textsuperscript{\rm 2} Alibaba Group \\
    \footnotesize\texttt{\{lsqlsq,ningci5252,gjn0310,yanyichao,zhuwenhan823,xkyang\}@sjtu.edu.cn}\\
    \footnotesize\texttt{jiangjing.ljj@alibaba-inc.com}
}

\begin{document}
\maketitle
\begin{abstract}
Texture editing is a crucial task in 3D modeling that allows users to automatically manipulate the surface materials of 3D models. 
However, the inherent complexity of 3D models and the ambiguous text description lead to the challenge in this task.
To address this challenge, we propose ITEM3D, a \textbf{T}exture \textbf{E}diting \textbf{M}odel designed for automatic \textbf{3D} object editing according to the text \textbf{I}nstructions.
Leveraging the diffusion models and the differentiable rendering, ITEM3D takes the rendered images as the bridge of text and 3D representation, and further optimizes the disentangled texture and environment map.
Previous methods adopted the absolute editing direction namely score distillation sampling (SDS) as the optimization objective, which unfortunately results in the noisy appearance and text inconsistency.
To solve the problem caused by the ambiguous text, we introduce a relative editing direction, an optimization objective defined by the noise difference between the source and target texts, to release the semantic ambiguity between the texts and images.
Additionally, we gradually adjust the direction during optimization to further address the unexpected deviation in the texture domain. 
Qualitative and quantitative experiments show that our ITEM3D outperforms the state-of-the-art methods on various 3D objects. We also perform text-guided relighting to show explicit control over lighting.
Our project page: \href{https://shengqiliu1.github.io/ITEM3D}{https://shengqiliu1.github.io/ITEM3D}.
\end{abstract}    

\section{Introduction}
\label{sec:intro}

Texture editing is an important task in 3D modeling that involves manipulating the surface properties of 3D models to create a visually appealing appearance according to the user's ideas. 
With the increasing applications of 3D models in entertainment and e-shopping, 
how to automatically generate and edit the texture of a 3D model without manual effort becomes an appealing task in the field of 3D vision.
However, this task is challenging due to the complexity of 3D models and the special representation of the texture. 


Recent advances have demonstrated the effectiveness of generative models in synthesizing high-quality appearances that are both visually pleasing and semantically meaningful. The use of generative adversarial networks (GANs)~\cite{DBLP:conf/cvpr/XianSARLFYH18,DBLP:conf/icml/BergmannJV17,DBLP:journals/tog/SunWSWWL22, chen2023hyperstyle3d} has shown promising results in producing textures with intricate patterns and complex structures. Other approaches, such as texture synthesis via direct optimization~\cite{DBLP:conf/iccv/EfrosL99,DBLP:journals/tog/FruhstuckAW19,DBLP:journals/tog/SendikC17a,DBLP:journals/tog/ZhouZBLC018} or neural style transfer~\cite{DBLP:journals/cgf/BerkitenHSMLR17,DBLP:journals/tog/HertzHGC20,DBLP:conf/aaai/WangZ0LZXL22,DBLP:conf/rt/MertensKCBD06}, have also been explored for their ability to generate textures with specific artistic styles. 
However, the capacity of these models is still unable to meet the need of real-world applications, which requires high-quality and diverse textures. 
Meanwhile, recent researches~\cite{chen2023fantasia3d,richardson2023texture,DBLP:journals/corr/abs-2211-07600,DBLP:journals/corr/abs-2211-10440, liu2023zero, kawar2022imagic, meng2021sdedit} on the diffusion models have emerged as a powerful new family of generative methods, which achieve impressive generation results in natural images and videos, inspiring us to introduce the awesome power into the task of 3D modeling.

\begin{figure}[t]
  \centering
  \includegraphics[width=\linewidth]{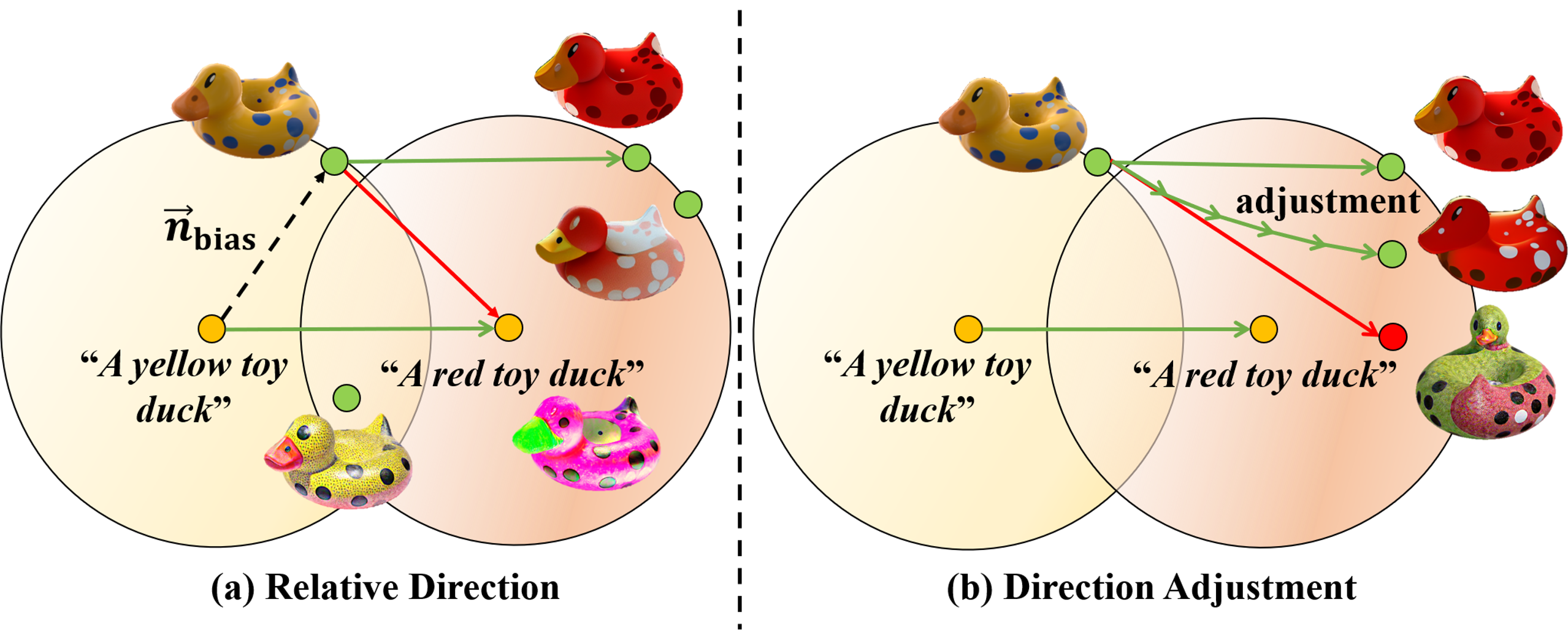}
   \caption{\textbf{Motivation}. (a) Previous methods~\cite{DBLP:journals/corr/abs-2209-14988, chen2023fantasia3d} with SDS loss to directly guide the optimization leads to ambiguous details due to the bias between texts and images (red line), while our method introduces the relative direction between source and target texts to the optimization process, eliminating the bias and improving the rendering results (green line). (b) The optimization in the texture domain gives rise to the deviation of the target direction (red line), thus we gradually adjust the direction to fine-tune the optimization (fold green line).}
   \label{fig:intro_1}
\end{figure}

However, directly applying the diffusion model to 3D objects is a non-trivial task due to the following reasons.
1) \textbf{The gap between the 3D representation and natural images.} Existing diffusion models are typically trained with natural images, making the pre-trained diffusion model lack prior knowledge in the 3D domain.
Moreover, due to the complexity of the 3D model, it would be difficult to jointly edit shape, appearance, and shading, sometimes leading to conflicts in optimization goals. Therefore, directly editing the 3D representation may cause extreme semantic bias and sacrifice inherent 3D topology.
2) \textbf{The learning misdirection of text description.} It is hard for text prompts to exactly describe the target images at the pixel level, leading to an ambiguous direction when taking the rendered images as the bridge.

To solve these problems, we present an efficient model, dubbed \textbf{ITEM3D}, which can generate visually natural texture corresponding to the text prompt provided by users.
Instead of directly applying the diffusion model to texture editing in the 2D space, we adopt rendered images as the intermediary that bridges the text prompts and the appearance of 3D models. 
Apart from the appearance, the lighting and shading are also key components influencing the rendering results.
Therefore, we represent the 3D model into a triangular mesh, a set of disentangled materials consisting of the texture, and an environment map following nvdiffrec~\cite{munkberg2022extracting}, which accomplishes the disentangled modeling of both appearance and shading.

To optimize a texture and an environment map with the diffusion model, a naive idea is to adopt the score distillation sampling (SDS)~\cite{DBLP:journals/corr/abs-2209-14988} like previous diffusion-based editing methods, which represents the absolute direction.
Unfortunately, the utilization of absolute direction, such as SDS, often leads to noisy details and an inconsistent appearance, due to the ambiguous description of the text prompt for the target images.
Inspired by the recent improvement~\cite{DBLP:journals/corr/abs-2304-07090}, we replace the absolute editing direction led by the score distillation sampling with a relative editing direction determined by two predicted noises under the condition of the source text and the target text respectively, as illustrated in~\cref{fig:intro_1}~(a). 
In this way, our model enables us to edit the texture in obedience to the text while bypassing the inconsistency problem by releasing the ambiguous description.
It is ideal that the intermediate states between the source and target text can give relatively accurate descriptions for arbitrary rendered images during the optimization, like the green straight lines in~\cref{fig:intro_1}~(b).
However, the optimization in the texture domain actually shows the nonlinear distortion of the appearance in rendered images, leading to the deviation from the determined direction, like the red line in~\cref{fig:intro_1}~(b).
To reduce the deviation caused by the texture projection, we gradually adjust the editing direction during the optimization, as green fold lines shown in~\cref{fig:intro_1}~(b).
With the advent of the textural-inversion model, it can be easy to automatically correct the description as the change of the texture and its rendered images.

Thanks to the proposed solutions, our method overcomes the challenges of domain gap and learning misdirection, fulfilling the requirements of texture editing.
In summary, our contributions are:
\begin{itemize}
\setlength\itemsep{0em}
\item We design an efficient optimization pipeline to edit the texture and environment map obedient to the text instruction, directly empowering the downstream application in the industrial pipeline.
\item We introduce the relative direction to the optimization of the textured 3D model, releasing the problem of noisy details and inconsistent appearance caused by the semantic ambiguity between the texts and images.
\item We propose to gradually adjust the relative direction guided by the source and target texts which addresses the unexpected deviation from the determined direction caused by the nonlinear texture projection.
\end{itemize}


\section{Related Work}
\label{sec:related}

\subsection{3D Model Representation}
\noindent\textbf{Neural Radiance Field.}
Neural implicit representations employ differentiable rendering techniques to reconstruct 3D geometry along with its corresponding appearance from multi-view images.
Neural Radiance Field (NeRF)~\cite{mildenhall2020nerf} and followup  approaches~\cite{zhang2020nerf++, wang2021nerf, muller2022instant, chen2022tensorf, wizadwongsa2021nex, yu2021plenoctrees, reiser2021kilonerf, martin2020nerf} utilize volumetric representations and a neural encoder to compute radiance field.
These methods explicitly represent the specular reflection as the appearance embedding on the surface, enabling the high-fidelity rendering results.
While these approaches are capable of capturing complex view-dependent appearances, they often suffer from poor surface reconstruction and limited relighting capabilities.
To maintain the geometry quality, surface-based methods~\cite{oechsle2021unisurf, wang2021neus, fu2022geo} introduce an implicit surface to the volumetric representation, resulting in improved geometry compared to pure NeRF-based methods.
Nevertheless, these representations require excessive computation resources and still face challenges in 
decoupling material components, lack of editability.

\noindent\textbf{Neural Inverse Rendering.}
To solve the aforementioned problem, another category of methods focuses on the decomposition of rendering parameters.
These methods introduce inverse rendering to estimate surface, materials, and lighting conditions simultaneously, achieving material editing and scene relighting.
NeRV~\cite{srinivasan2021nerv} and NeRD~\cite{boss2021nerd} decompose the scene into the shape, reflectance and illumination components, enabling rendering under varying lighting conditions.
NeRFactor~\cite{zhang2021nerfactor} introduces a two-stage method to separately reconstruct geometry and reflectance, simplifying reflectance optimization.
Nvdiffrec~\cite{munkberg2022extracting} additionally introduces a differentiable model of environment lighting to efficiently recover all-frequency lighting.
Considering the balance between appearance and editability, our proposed method, ITEM3D, leverages this decomposed representation to enable efficient and natural texture editing while preserving the topology of objects.


\subsection{3D Text-guided Generation}
\noindent\textbf{CLIP-based Generation.}
With the advent of large text-image models, CLIP, recent works~\cite{wang2022clip, jetchev2021clipmatrix, jain2022zero, michel2022text2mesh} have led to impressive progress in 3D text-driven synthesis.
The majority of methods adopt optimization procedures supervised by the CLIP similarity~\cite{radford2021learning}.
Specifically, CLIP-NeRF~\cite{wang2022clip} proposes a unified framework to manipulate NeRF, guided by either a text prompt or an example image.
Similarly, CLIP-Mesh utilizes an explicit textured mesh as a 3D representation, allowing for shape deformation along with corresponding texture modifications based on the input text.
As the 3D-aware GAN gains popular, several works~\cite{gal2022stylegan, chen2023hyperstyle3d, kim2023datid} focus on the 3D stylization by fine-tuning the 3D GAN under the guidance of the CLIP model.
However, the main drawback of CLIP-guided generation is the lack of diversity due to the deterministic embedding of the target prompt.

\noindent\textbf{Diffusion-based Generation.}
Apart from the CLIP-based method, the diffusion model~\cite{ho2020denoising} has recently inspired huge breakthroughs in 3D text-guided generation.
These methods can be briefly classified into two categories, \ie, 3D large generative models~\cite{liu2023zero,chen2023single,shue20233d} and optimization for specific 3D representations~\cite{haque2023instruct, DBLP:journals/corr/abs-2209-14988, chen2023fantasia3d, tang2023make, zeng2023avatarbooth, cao2023dreamavatar,DBLP:journals/corr/abs-2211-07600}.
Large models rely on large-scale 3D data to train a diffusion model aware to 3D objects.
One of these related methods,
Point-UV diffusion~\cite{yu2023texture}, proposes a two-stage architecture containing a point diffusion and a UV diffusion to generate textures for 3D meshes.
Despite the fast inference, these methods faces the challenges of low quality and poor generalization for the complex objects, due to the amount and diversity of current 3D datasets.

Considering the limitation of data, more researchers focus on the optimization of 3D representation for the specific objects.
Dreamfusion~\cite{DBLP:journals/corr/abs-2209-14988} initially proposes the SDS loss as a means to incorporate the 2D prior of Stable Diffusion~\cite{rombach2022high} into the generation of 3D domains.
Following Dreamfusion, several works~\cite{tang2023make, zeng2023avatarbooth, cao2023dreamavatar,DBLP:journals/corr/abs-2211-07600} incorporate the SDS loss to directly optimize NeRF-like representations.
Different from the NeRF representations, DreamGaussian~\cite{tang2023dreamgaussian} adopts the recently advent Gaussian splatting as the 3D representation and introduces the 3D pre-trained diffusion model~\cite{liu2023zero} to provide the additional geometry prior.
Fantasia3D~\cite{chen2023fantasia3d} leverages the disentangled modeling, and learns the geometry and appearance supervised by the score distillation sampling.
However, these SDS-based methods often produce non-detailed and blurry outputs due to noisy gradients.
In contrast, our ITEM3D utilizes the relative direction to eliminate the semantic ambiguity of the target prompt towards the texture.
Recently, several works~\cite{haque2023instruct,chen2023text2tex,raj2023dreambooth3d} utilize 2D diffusion model~\cite{brooks2023instructpix2pix, zhang2023adding} to generate the initial image guidance and alternatively refine the 3D model.
Instruct-NeRF2NeRF~\cite{haque2023instruct} introduces the NeRF optimization with training images that are gradually updated by the Instruct-Pix2Pix~\cite{brooks2023instructpix2pix}.
While their model can achieve high-quality rendering results, the training process is too lengthy as they need an updating strategy to maintain the 3D consistency of NeRF representation during the editing process.
Similarly, Text2Tex~\cite{chen2023text2tex} initializes one-view appearance for the given mesh by the Controlnet~\cite{zhang2023adding} and progressively generates the novel-view texture while refining the existed texture.
In contrast, the decomposed 3D representation of our ITEM3D empowers us to directly optimize the disentangled materials, texture map and environment map, which significantly improves the training efficiency.

\section{Method}

\begin{figure*}[t]
  \centering
  \includegraphics[width=\linewidth]{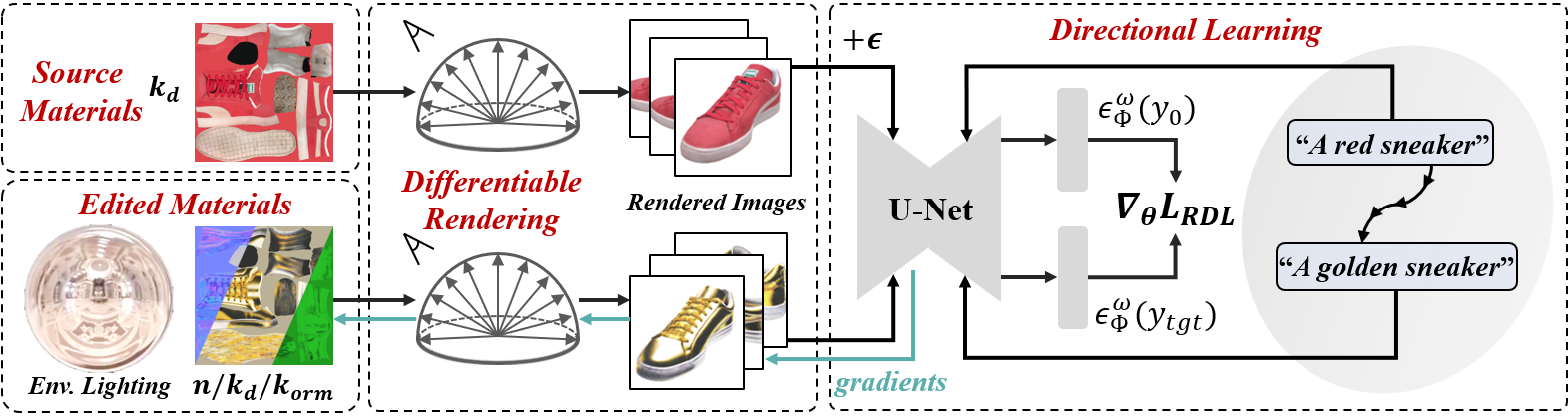}
   \caption{\textbf{Pipeline} of 3D model texture editing. We render the 3D model with mesh, texture, and environment map into 2D images which are then added with noise $\epsilon$. We further separately use the target text and the gradually adjusted source text as the conditions to predict the added noise via the U-Net. The difference between the two predicted noises serve as the relative direction to guide the optimization of the materials and environment map.}
   \label{fig:method_1}
\end{figure*}

\subsection{Overview}
Given a set of multi-view images $\mathcal{I}=\{I_{1}, ..., I_{n}\}$, we first reconstruct the 3D model with both geometry and texture, and then edit the texture under the guidance of text prompts.
To this end, we design a zero-shot differentiable framework that optimizes the disentangled materials of the object, \ie, texture map and environment map.
We leverage a differentiable rendering model to represent the 3D model as an accurate shape and surface materials with texture and environment map (\cref{sec:Representation}).
For further editing of appearance, we utilize the diffusion model to guide the direction of the texture optimization given the target text prompt.
To solve the problem of ambiguous and noisy details, we introduce the relative direction of source texts and target texts into the optimization (\cref{sec:Relative direction}). 
Moreover, we gradually adjust relative direction to address the challenges of deviation caused by the unbalanced optimization in the texture domain (\cref{sec:Adjustment}).
The overview of our method is demonstrated in~\cref{fig:method_1}.

\subsection{3D Model Representation}
\label{sec:Representation}
To accomplish disentangled editing, we decompose the 3D model into a triangular mesh, a set of spatially varying materials and an environment map.
The material model we employ, denoted as $(k_d, k_{orm}, n)$, combines a diffuse term $k_d$, a specular term $k_{orm}$, and a normal term $n$ following the physically-based (PBR) material model introduced by Disney~\cite{burley2012physically}.
This PBR material model can be seamlessly integrated into existing industry rendering engines without any change.

Additionally, for efficient representation of volumetric textures, we leverage a multi-layer perceptron (MLP) to encode all material parameters into a compact representation.
Specifically, given a world space position $x$, we compute the MLP texture, which includes the base color, $k_d$, the specular parameters $k_{orm}$, and a tangent space normal perturbation $n$. This mapping is formulated as $x \rightarrow (k_d, k_{orm}, n)$.
With the introduction of neural texture representation, the textures are initialized by sampling the MLP on the mesh surface and then optimized efficiently, while maintaining a fixed topology.

Following the rendering equation of the image-based lighting model, we integrate all the incoming radiance $L_i\left(\omega_i\right)$ from the environment map and compute the outgoing radiance $L$ in direction $\omega_o$ using the following equation:
\begin{equation}
L\left(\omega_o\right)=\int_{\Omega} L_i\left(\omega_i\right) f\left(\omega_i, \omega_o\right)\left(\omega_i \cdot \mathbf{n}\right)d\omega_i ,
\label{eq:render}
\end{equation}
where the $f\left(\omega_i, \omega_o\right)$ represents the BSDF function, and the integration domain is the hemisphere $\Omega$ around the intersection normal $n$.
For real-time rendering, we employ the split-sum approximation~\cite{karis2013real} and the integral radiance in ~\cref{eq:render} is approximated as:
\begin{equation}
\begin{split}
    L\left(\omega_o\right) \approx &\int_{\Omega} f\left(\omega_i, \omega_o\right)\left(\omega_i \cdot \mathbf{n}\right) d \omega_i \\
&\int_{\Omega} L_i\left(\omega_i\right) D\left(\omega_i, \omega_o\right)\left(\omega_i \cdot \mathbf{n}\right) d \omega_i.
\end{split}
\end{equation}
The first term of this product only relies on the parameters $\left(\omega_i \cdot \mathbf{n}\right)$ and the roughness $r$ of the BSDF $f\left(\omega_i, \omega_o\right)$, which is precomputed and stored in a 2D lookup texture. Meanwhile, the second term is the integral of the incident radiance with the GGX~\cite{DBLP:conf/rt/WalterMLT07} normal distribution $D$, which is also precomputed and stored by a filtered cubemap following Karis~\cite{karis2013real}.

In order to learn the environment lighting from 2D image observations, we employ a differentiable split sum shading model. We represent the environment light as a high-resolution cube map, which is the trainable parameters in the shading model.
Owing to the precomputation and lookup mechanism, the rendering process from the texture/normal/environment map to the 2D images can be greatly accelerated.
This approach serves as a bridge that allows ITEM3D to directly optimize the texture and environment map through rendered 2D images, rather than optimizing the complex 3D representation. As a result, the number of parameters to be optimized is significantly reduced, leading to a more efficient editing process.

\subsection{Relative Direction Based Optimization}
\label{sec:Relative direction}
To enable the appearance editing of 3D models using natural language, 
a direct idea is to utilize the diffusion model that has been pre-trained in 2D images as knowledge prior.
Naively, we can use Score Distillation Sampling (SDS) loss,
\begin{equation}
\nabla_{\theta}\mathcal{L}_{\mathrm{SDS}}(\phi, \mathbf{x}=\mathcal{R}(\theta))=\mathbb{E}_{t,\epsilon}\left[w(t)\left(\epsilon_{\phi}^{\omega}(\mathbf{z}_t;y,t)-\epsilon\right)\frac{\partial\mathbf{x}}{\partial\theta}\right],
\end{equation}
where $\mathbf{x}$ is the rendered images, $t$ is the sampled time step, $z_t$ is the $t$ time step latent, $w(t)$ is the weighting function that equals $\partial\mathbf{z}_t / \partial\mathbf{x}$, $y$ is the text condition, $\epsilon_{\phi}^{\omega}(\mathbf{z}_t;y,t)$ is the predicted noise through classifier-free guidance, and $\epsilon \in N (0, I)$ is the noise added to the rendered images.
The gradient of SDS loss gives an editing direction for our texture optimization, determined by the text prompt $y$.
However, the SDS loss may harm the content of original images with noisy details, because the text prior typically cannot faithfully reflect the information of the image.
It is known that the entropy of an RGB image is significantly larger than that of a text prompt.
As a consequence, the misdescription inevitably arises when taking the text prompt as the prior to restore the high-quality image from the same-scale noise.
Therefore, even for a text prompt $y_{0}$ describing the original images, there exists a deviation related to the optimized texture $\theta$ between the added noise $\epsilon$ and the predicted noise $\epsilon_{\phi}^{\omega}\left(\mathbf{z}_{t};y_{0},t\right)$, which can be simply expressed as, 
\begin{equation}
    D_{\mathrm{bias}}(\theta, ... ) \propto ||\epsilon_{\phi}^{\omega}\left(\mathbf{z}_{t};y_{0},t\right)-\epsilon||.
\end{equation}

Thus, the gradient leads to a bias term from the original input image, which can be expressed as, 
\begin{equation}
    \Vec{n}_{\rm bias} = \frac{\partial D_{\mathrm{bias}}(\theta, ...)}{\partial\theta}  = \left(\epsilon_{\phi}^{\omega}\left(\mathbf{z}_{t};y_{0},t\right)-\epsilon\right)\frac{\partial\mathbf{x}}{\partial\theta}.
\end{equation}
Moreover, for an arbitrary text prompt $y_{\rm tgt}$ describing the target editing texture, it could be considered that there exists a term of expected editing direction and a term of bias discussed above, 
\begin{equation}
\left(\epsilon_{\phi}^{\omega}\left(\mathbf{z}_{t};y_{\rm tgt},t\right)-\epsilon\right)\frac{\partial\mathbf{x}}{\partial\theta} = \Vec{n}_{\rm tgt} + \Vec{n}_{\rm bias}.
\end{equation}
As a result, the $\Vec{n}_{\rm bias}$ gives rise to the misdirection for the optimization procedure.

To address these issues, it is ideal to find the accurate editing direction $\Vec{n}_{\rm tgt}$, while the term of $\Vec{n}_{\rm bias}$ is hard to estimate due to the diverse input images.
To mitigate the gap, it is natural to take the text guidance as a relative direction rather than an absolute direction, enabling us to eliminate the term of $\Vec{n}_{\rm bias}$.
The absolute direction of the source $\Vec{n}_{\rm src}$ and the target $\Vec{n}_{\rm tgt}$ can be expressed as,
\begin{equation}
    \Vec{n}_{\rm src} = \left(\epsilon_{\phi}^{\omega}\left(\mathbf{z}_{t};y_{0},t\right)-\epsilon\right)\frac{\partial\mathbf{x}}{\partial\theta} - \Vec{n}_{\rm bias} ,
\end{equation}
\begin{equation}
    \Vec{n}_{\rm tgt} = \left(\epsilon_{\phi}^{\omega}\left(\mathbf{z}_{t};y_{\rm tgt},t\right)-\epsilon\right)\frac{\partial\mathbf{x}}{\partial\theta} - \Vec{n}_{\rm bias},
\end{equation}
where $\Vec{n}_{\rm src}$ is actually the $\Vec{0}$ giving no extra information to the input images.
Inspired by the CLIP-directional loss improved by the StyleGAN-Nada~\cite{DBLP:journals/tog/GalPMBCC22} and the denoising loss proposed by the recent work~\cite{DBLP:journals/corr/abs-2209-14988}, we utilize the difference between the source $\Vec{n}_{\rm src}$ and the target $\Vec{n}_{\rm tgt}$ as the relative direction of the target, which can be presented as, 
\begin{equation}
    \Vec{n}_{\rm tgt}  = \Vec{n}_{\rm tgt} - \Vec{n}_{\rm src} = \left(\epsilon_{\phi}^{\omega}\left(\mathbf{z}_{t},y_{\rm tgt},t\right)-\epsilon_{\phi}^{\omega}\left(\mathbf{x},y_{0},t\right)\right)\frac{\partial\mathbf{x}}{\partial\theta}.
\end{equation}
Therefore, the final gradient utilized for optimizing the texture can be presented as,
\begin{equation}
    \begin{split}
&\nabla_{\theta}\mathcal{L}_{\mathrm{RDL}}(\phi, \mathbf{x}=\mathcal{R}(\theta))= \\ &\mathbb{E}_{t,\epsilon}\left[w(t)\left(\epsilon_{\phi}^{\omega}(\mathbf{z}_t;y_{\rm tgt},t)-\epsilon_{\phi}^{\omega}(\mathbf{z}_t;y_{0},t)\right)\frac{\partial\mathbf{x}}{\partial\theta}\right].
    \end{split}
\end{equation}

\subsection{Direction Adjustment}
\label{sec:Adjustment}
Different from the gradual transition in the nature image domain, the optimization of the texture domain unfortunately shows an unexpected offset of the appearance in rendered images.
The inherent reason is that the complexity of rendering leads to unbalanced optimization for the texture, with some parts under-tuning and other parts over-tuning.
This appearance offset can be seen in some parts of the rendered image, leading to the inconsistency between the source text and the rendered images in the median period of the optimization procedure.
It is known that a source image with an inconsistent text description means an optimization misdirection which leads to an unknown change in the editing results.
Similarly, if a rendered image during the intermediate optimization hops out of the direction between the source text and the target text, it can be considered as an inconsistent description of the source image when we take the current median point as a relative beginning point.
The original editing direction is given by, 
\begin{equation}
    \Vec{n}_{\rm ori} = \Vec{n}_{\rm tgt} - \Vec{n}_{\rm src}.
\end{equation}
If the optimization continues along the original direction, a more severe deviation of the material will be attached to the editing procedure.

To avoid the misdirection caused by the texture domain, we propose to adjust the editing direction, gradually changing the source text prompt during our optimization process of the texture map. The direction after adjustment $\mathrm{\Delta}\hat{T}_{i}$ can be represented as, 
\begin{equation}
    \mathrm{\Delta}\hat{T}_{i} = \Vec{\hat{n}}_{i} - \Vec{\hat{n}}_{i-1} = \mathcal{B}(I_{i}) - \mathcal{B}(I_{i-1}),
\end{equation}
where $i$ is the optimization iteration and $\mathcal{B}\left(\bm{\cdot} \right)$ expresses the inverse text generated by a pre-trained language-image model BLIP-v2~\cite{li2023blip}. As shown in the~\cref{fig:intro_1}~(b), the direction is continually adjusted during the optimization so that the new global direction $\Vec{n}_{i}$ can be written as, 
\begin{equation}
    \Vec{n}_{i} = \Vec{n}_{\rm ori} + \sum_{j \leq i}{ \mathrm{\Delta}\hat{T}_{j}}.
\end{equation}
By adjusting the optimization direction step by step, 
we achieve more delicate and controllable editing, which can be seen in~\cref{sec:dir_adjust}.

\section{Experiments}

\begin{figure*}[t]
  \centering
  \includegraphics[width=\linewidth]{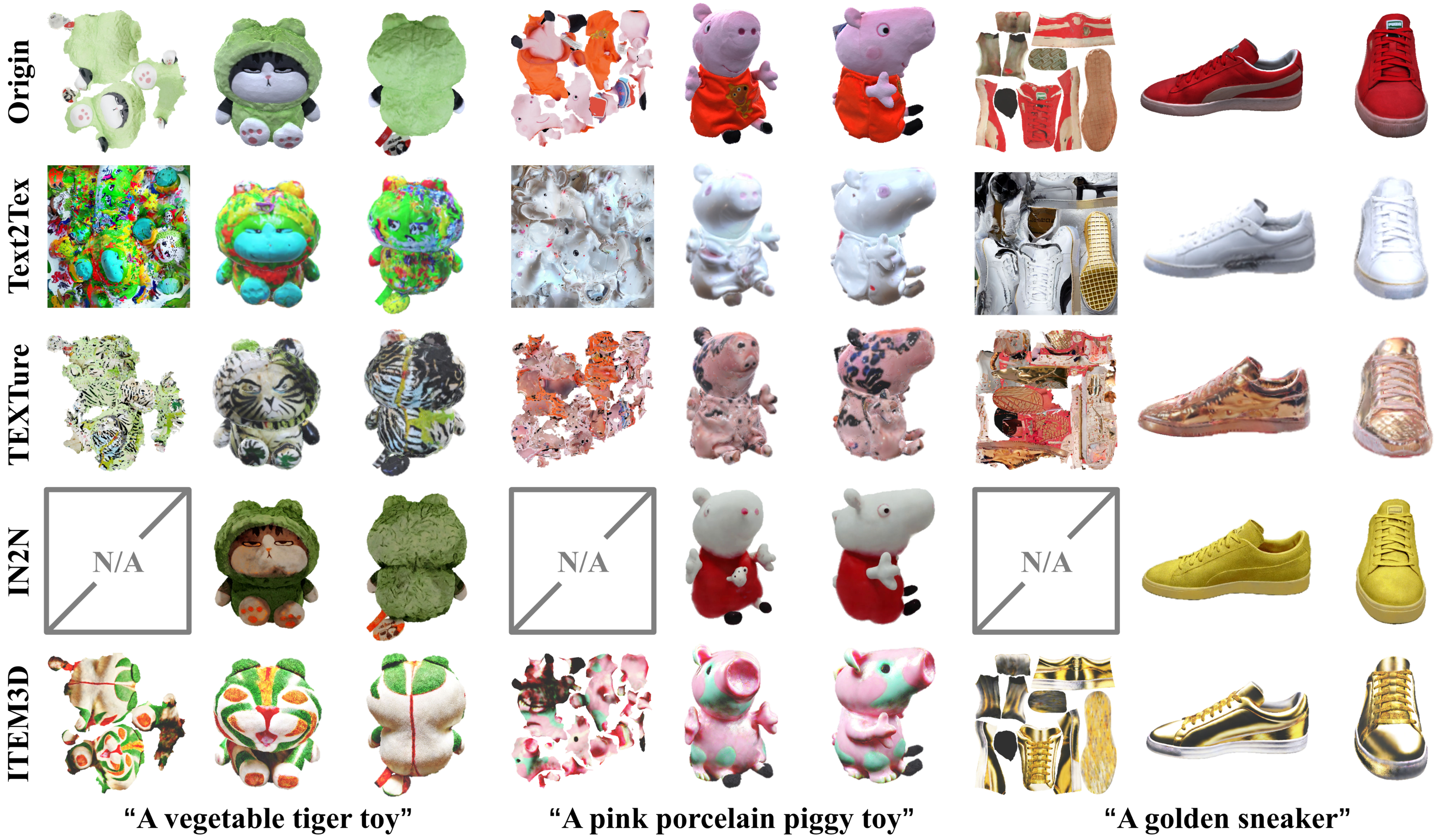}
   \caption{\textbf{Performance on real-world objects.} ITEM3D successfully transform the original cat toy into a vegetable tiger toy, the piggy doll into a porcelain pig, and the sneaker into a golden sneaker with remarkable quality. In contrast, Text2Tex~\cite{chen2023text2tex} and TEXTure~\cite{richardson2023texture} both generate noisy textures resulting in low-quality rendering appearance. Instruct-NeRF2NeRF~\cite{haque2023instruct} achieves natural and text-consistent appearance, but it fails to edit the material of the objects. It is noted that the text instructions given to the Instruct-NeRF2NeRF are slightly different from other methods, \ie, \textit{``Make it into a vegetable tiger toy''}, \textit{``Make it into a porcelain piggy toy''} and \textit{``Make it into a golden sneaker''}, due to its special requirement.}
   \label{fig:real_dataset}
\end{figure*}

\subsection{Experiment Setup}
\noindent\textbf{Dataset.}  In the experiments, we evaluate our model on the NeRF Synthetic~\cite{mildenhall2020nerf} dataset, Keenan's 3D model repository and a set of real-world products.
The NeRF synthetic dataset consists of 8 path-traced scenes with multi-view images which we reconstruct into our textural mesh-based representation via nvdiffrec~\cite{munkberg2022extracting}.
The Keenan's 3D model repository and real-world dataset contains explicit mesh and texture map for each 3D model.

\noindent\textbf{Implementation Details.} We optimize 3D models on one RTX A6000 GPU with 48G memory. The procedure lasts about 600 iterations within 10 minutes for each 3D model. 

\begin{figure*}[t]
  \centering
  \includegraphics[width=\linewidth]{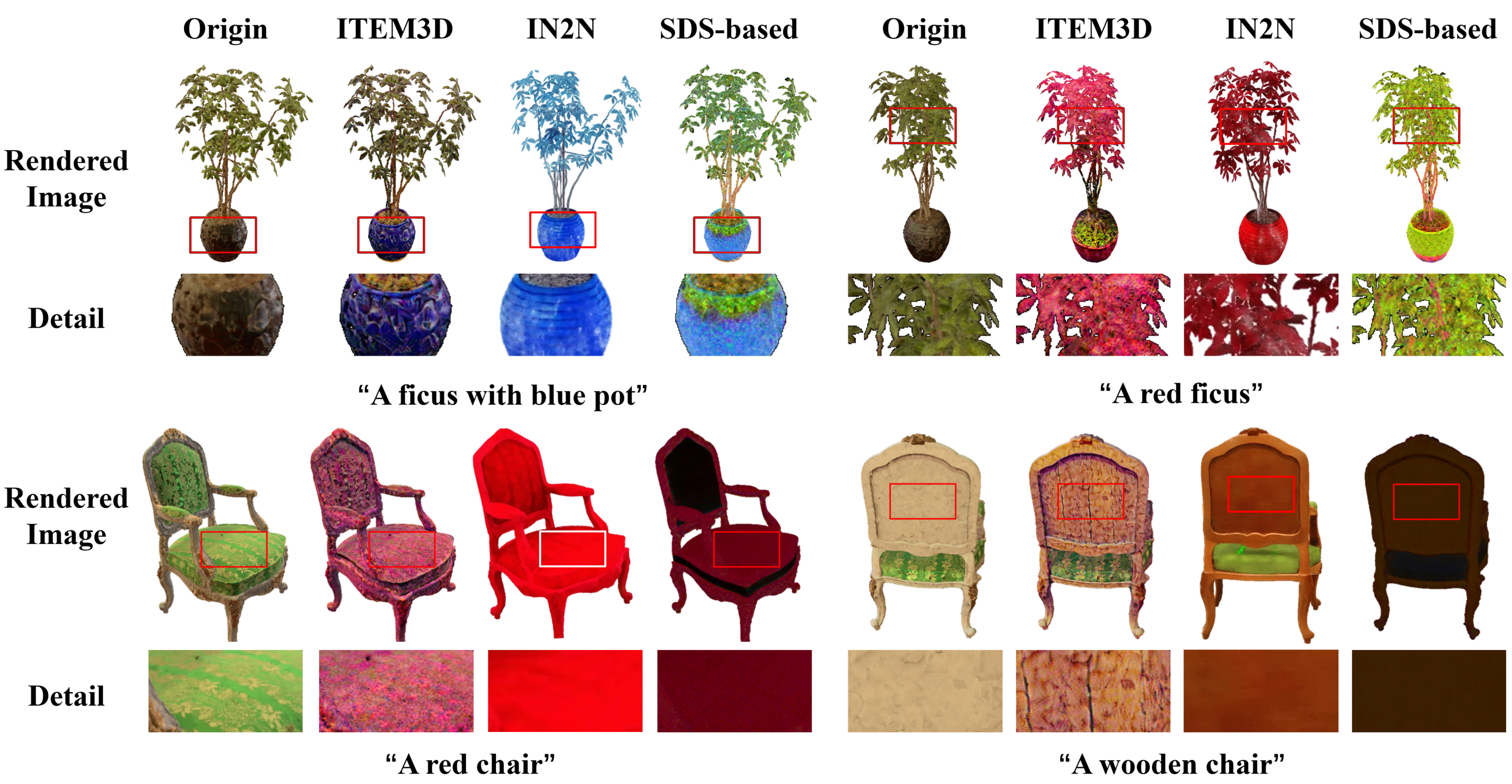}
   \caption{\textbf{Qualitative comparison} on NeRF synthetic dataset. We conducted an analysis of our method with the state-of-the-art approach, Instruct-NeRF2NeRF (IN2N)~\cite{haque2023instruct} and the simple SDS-based method. While both ITEM3D and IN2N demonstrate prompt-consistent editing, the SDS-based method fails to yield satisfactory outcomes. Conversely, IN2N exhibits a loss of the original chair patterns and an inability to faithfully represent natural wood patterns. Moreover, it lacks the necessary precision to accurately discern the edited object.}
   \label{fig:exp_1}
\end{figure*}

\subsection{Qualitative Results}
\noindent\textbf{Results on Real-world Objects.} We perform our ITEM3D, TEXTure~\cite{richardson2023texture}~(SIGGRAPH 2023), Text2Tex~\cite{chen2023text2tex}~(ICCV 2023) and Instruct-NeRF2NeRF~\cite{haque2023instruct}~(ICCV 2023) on a real-world dataset, and the results are depicted in \cref{fig:real_dataset}.
Our ITEM3D showcases remarkable texture editing capabilities for real products, including a shoe, a piggy doll, and a toy cat.
For instance, when given the text prompt \textit{``A vegetable tiger toy''}, ITEM3D effectively transforms the texture into that of a cute, furry tiger while preserving the structure of the original toy cat. 
Furthermore, when the prompts involve both material and texture descriptions, \eg, \textit{``A golden sneaker''}, our model successfully bakes the golden material instead of the original material into the texture and shows a realistic shoe.
Similarly, the example of \textit{``A pink porcelain piggy toy''} also
achieves realistic material editing along with consistent texture modifications as the provided text.
Compared to our methods, Instruct-NeRF2NeRF~\cite{haque2023instruct} fails to learn the materials of porcelain and gold in the cases of the piggy toy and sneaker despite the natural appearance.
Meanwhile, it is difficult for Text2Tex~\cite{chen2023text2tex} and TEXTure~\cite{richardson2023texture} to handle complex objects, resulting in low-quality appearances.
Besides, Text2Tex suffers from the problem of multiple heads in the example of tiger toy, due to its patch-based texture learning.
These results clearly demonstrate the generalization ability of ITEM3D in handling complex real-world objects.

\noindent\textbf{Comparison on Synthetic Objects.} We also compare ITEM3D with the SDS-based optimization method, and Instruct-NeRF2NeRF (IN2N)~\cite{haque2023instruct} in the NeRF synthetic dataset without explicit mesh and texture, as shown in~\cref{fig:exp_1}.
While the SDS-based method can edit textures along the direction of the text prompt, its rendered images show an unrealistic appearance and tend to overfit the text. In contrast, ITEM3D and IN2N can render realistic images with high quality while remaining consistent with the input text prompt. However, IN2N has certain limitations in comparison to our approach.
For instance, when editing chairs, IN2N produces results with smooth textures, while our method captures more fine-grained details. Furthermore, when editing a ficus plant with the prompt \textit{``Turn the pot into a blue pot''}, IN2N faces the challenge of distinguishing between the pot and the plant, resulting in both elements being edited to the same blue color.
The comparison indicates the effectiveness of the introduced relative direction and direction adjustment.

\subsection{Quantitative Results}

\begin{table*}[t]
  \centering
  \resizebox{\linewidth}{!}{
  \begin{tabular}{@{}cc|cccc|cc@{}}
    \toprule
    Method & Venue & \makecell{Global \\ Score} $\uparrow$ &\makecell{Directional \\ Score} $\uparrow$ & \makecell{Training \\ Time} $\downarrow$ & \makecell{GPU \\ Memory} $\downarrow$ & Photorealism $\uparrow$ & \makecell{Text \\ Consistency} $\uparrow$ \\
    \midrule
    Text2Tex~\cite{chen2023text2tex} & ICCV 2023 & 0.294 & 0.121  & 15 min & 10 GB & 3.62 & 2.79\\
    TEXTure~\cite{yu2023texture} & SIGGRAPH 2023 & 0.299 & 0.097 & \textbf{5 min} & 12 GB & 3.03 & 2.41\\
    IN2N~\cite{haque2023instruct} & ICCV 2023 & 0.311 & 0.159  & 10 h & 15 GB & 4.26 & 3.36\\
    ITEM3D (ours) & \textemdash & \textbf{0.333} & \textbf{0.174} & 8 min & \textbf{9 GB} & \textbf{4.38} & \textbf{4.18}\\
    \bottomrule
  \end{tabular}}
  \caption{\textbf{Quantitative Comparisons}. We report two CLIP-based scores, \ie, global score and directional score to evaluate the semantic quality of rendered images. We also compare the editing time and memory consumption of four methods. Besides, we conduct a user study with 33 participants. Each participant scores based on two evaluation criteria, \ie, photorealism and text consistency.}
  \label{tab:comparison}
\end{table*}

We perform a quantitative comparison with Text2Tex~\cite{chen2023text2tex}, TEXTure~\cite{richardson2023texture}, and IN2N~\cite{haque2023instruct} on the editing quality, efficiency and user study as presented in~\cref{tab:comparison}.

\noindent\textbf{Editing Quality.} To evaluate the semantic consistency, we render $512 \times 512$ RGB images after texture editing, and further compute the CLIP-Score of the rendered image and corresponding target text.
CLIP-score contains two parts, \ie, global score and directional score. Global score measures the similarity between the target text and the editing images, and directional score measures the similarity between two editing directions of text prompts and images, which can be expressed as,  
\begin{equation}
\text{Score}_{\rm global} = \frac{{T}_{\rm tgt} \cdot {I}_{\rm tgt}}{\lVert {{T}_{\rm tgt} \rVert \lVert {I}_{\rm tgt} \rVert}},
\quad
\text{Score}_{\rm direction} = \frac{\mathrm{\Delta}{T} \cdot\mathrm{\Delta}{I}}{\lVert \mathrm{\Delta}{T} \rVert \lVert \mathrm{\Delta}{T} \rVert},
\end{equation}
where ${T}_{\rm tgt}$ and ${I}_{\rm tgt}$ are the embedding of target text and edited image encoded by the CLIP encoder, and $\mathrm{\Delta}{T}$ and $\mathrm{\Delta}{I}$ are expressed as,
\begin{equation}
    \mathrm{\Delta}{T}={T}_{\rm tgt}-{T}_{\rm src},
    \quad
    \mathrm{\Delta}{I}={I}_{\rm tgt}-{I}_{\rm src}.
\end{equation}
As shown, our method achieves better results than Text2Tex, TEXTure and IN2N in both global score and directional score, demonstrating the superiority of our method in the image quality and text consistency.

\noindent\textbf{Editing Efficiency.} Moreover, to demonstrate the editing efficiency of ITEM3D, we also compare the training time and GPU memory with Text2Tex, TEXTure and IN2N. Our findings demonstrate that ITEM3D exhibits comparable time efficiency to TEXTure, while outperforming Text2Tex and IN2N in texture editing. Moreover, our method ITEM3D consumes less GPU memory compared to other approaches.

\noindent\textbf{User Study.} Additionally, we perform a user study to further assess the quality of editing objects. Users are required to rate on a scale of 1 to 5, based on the following questions:
(1) Are the edited objects realistic and natural (Photo-realism)? 
(2) Do the edited objects accurately reflect the target text’s semantics (Text Consistency)?
The results demonstrate the superior quality with higher realism and more text consistency of our proposed method as compared to the baselines.
The results of user study are also consistent to the qualitative analysis that IN2N achieves natural image appearance but fails on text consistency, while Text2Tex and TEXTure have lower image quality.


  

\begin{figure}[t]
  \centering
  \includegraphics[width=0.92\linewidth]{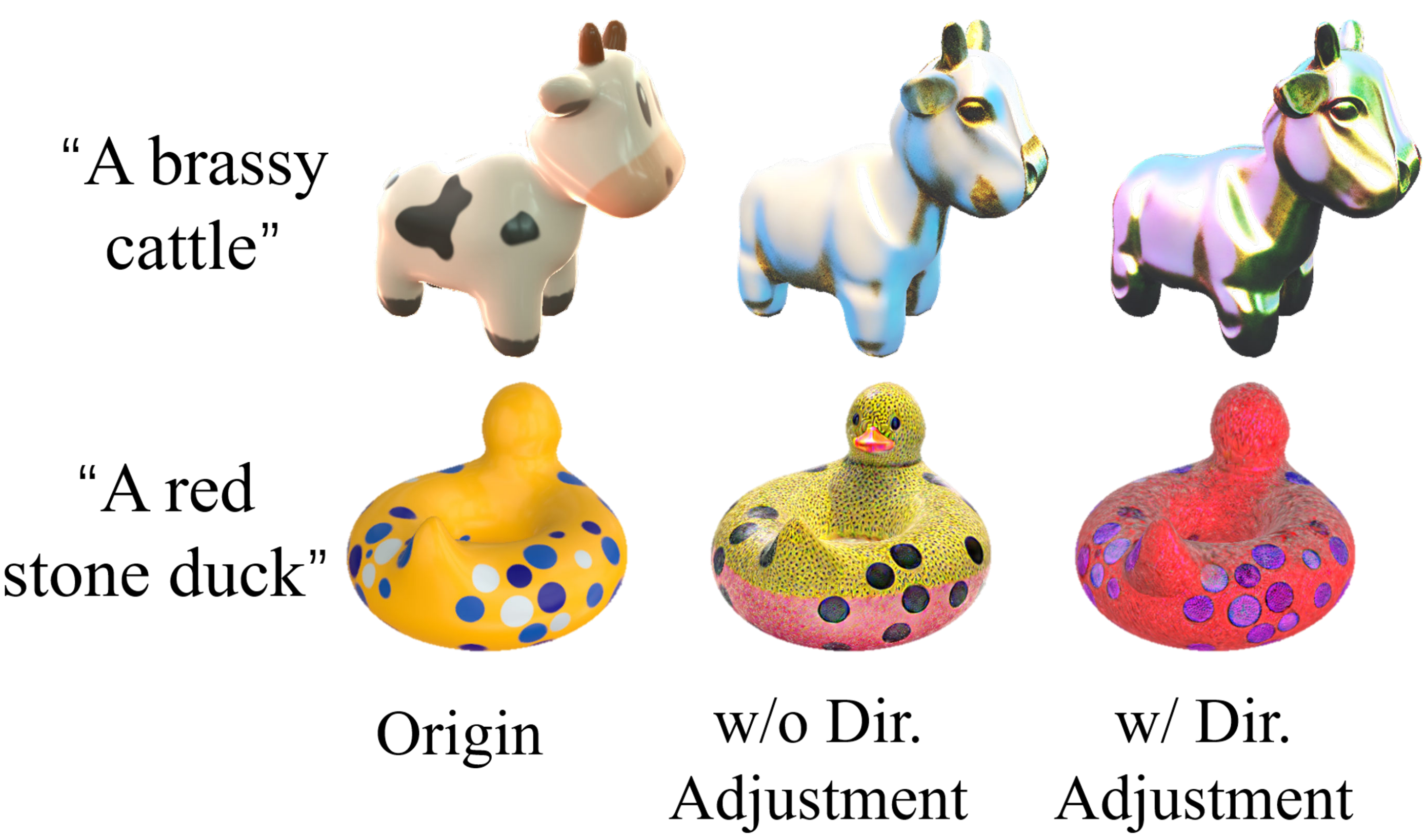}
   \caption{\textbf{Ablation study} of direction adjustment. The results without adjustment show a wired appearance, \ie, a pale body of the cattle and dual heads of the duck. However, with the application of gradual adjustment, the unrealistic artifacts are released, leading to a natural appearance.}
   \label{fig:exp_2}
\end{figure}

\subsection{Direction Adjustment}
\label{sec:dir_adjust}
In this section, we further study the necessity of direction adjustment. 
We perform the ablation study in~\cref{fig:exp_2}.
Without the adjustment for the relative optimization direction, the texture shows a wired change that the duck gradually generates two heads and the color seems partially yellow and partially red.
When applying the gradual adjustment, the duck bypasses the unnatural change and smoothly achieves the target appearance.
The example of cattle shows a similar trend.
In this experiment, it can be noticed that there exists unbalanced optimization for different parts of the texture.
The optimization scheme of simple pieces of texture converges quickly, while more complex modifications require longer time, which in turn over-tunes easy parts leading to poor results.

\begin{figure}[t]
  \centering
  \includegraphics[width=\linewidth]{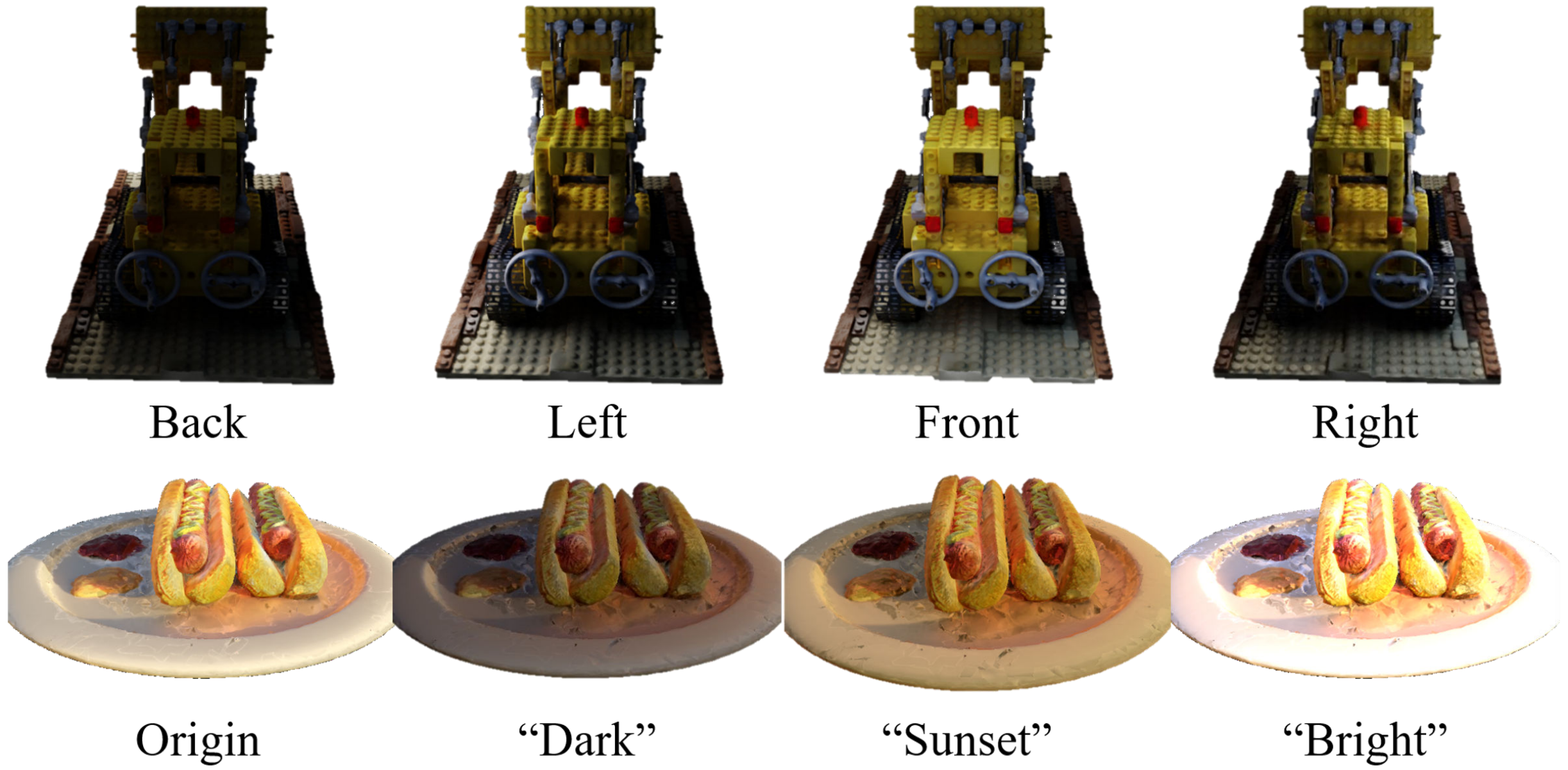}
   \caption{\textbf{Relighting results.} ITEM3D has capacity of explicit control over the lighting direction by manipulating the environment map. Furthermore, ITEM3D can relight the 3D model based on the text guidance via the optimization of the environment map.}
   \label{fig:relighting}
\end{figure}

\subsection{Relighting Application}
The disentangled representation of the environment map empowers ITEM3D to explicitly relight the 3D model.
The results of relighting are demonstrated in~\cref{fig:relighting}.
The Lego cases show the capability to modify the lighting direction by directly manipulating the environment map.
Moreover, the hotdog cases showcase the ability to guide relighting using text prompts, while maintaining the texture constant.
During the text-guided relighting, we fix the texture and mesh components and focus on optimizing the environment map based on provided text prompts, such as \textit{``dark''}, \textit{``sunset''} and \textit{``bright light''}.
It proves that the diffusion model can utilize the lighting prior information from 2D images to guide the optimization for the environment map by leveraging the differential rendering process.


\section{Conclusion and Limitations}
In conclusion, our ITEM3D model presents an efficient solution to the challenging task of texture editing for 3D models.
By leveraging the knowledge from diffusion models, ITEM3D is capable to optimize the texture and environment map under the guidance of text prompts.
To address the semantic ambiguity between text prompts and images, we replace the traditional score distillation sampling (SDS) with a relative editing direction. We further propose a gradual direction adjustment during the optimization procedure, solving the unbalanced optimization in the texture.

Despite the promising editing results, our ITEM3D still remains several limitations which should be solved in future work.
The major limitation is that our method may encounter challenges when dealing with complex and precise editing prompts.
Another limitation is that it is hard to edit the lighting direction directly by the guidance of text prompt.
Our further work aims to explore the UV mapping between neural texture and rendered images to achieve the disentangled and precise texture editing.

{
    \small
    \bibliographystyle{ieeenat_fullname}
    \bibliography{main}
}

\clearpage
\setcounter{page}{1}
\maketitlesupplementary
Although the main paper stands on its own, it is still worthwhile to provide more details.
In this supplementary document, we provide
\begin{itemize}
    \item More qualitative results and additional comparisons
    \item Failure cases in precise editing.
    \item Broader impacts of our ITEM3D.
\end{itemize}
The code of our framework used in our experiments will be made publicly available.
Additionally, we also provide a supplementary video that offers a brief introduction of our work and showcases 3D results rendered by a camera following a circular trajectory.

\section{Additional Qualitative Results}
\begin{figure*}[t]
  \centering
  \includegraphics[width=\linewidth]{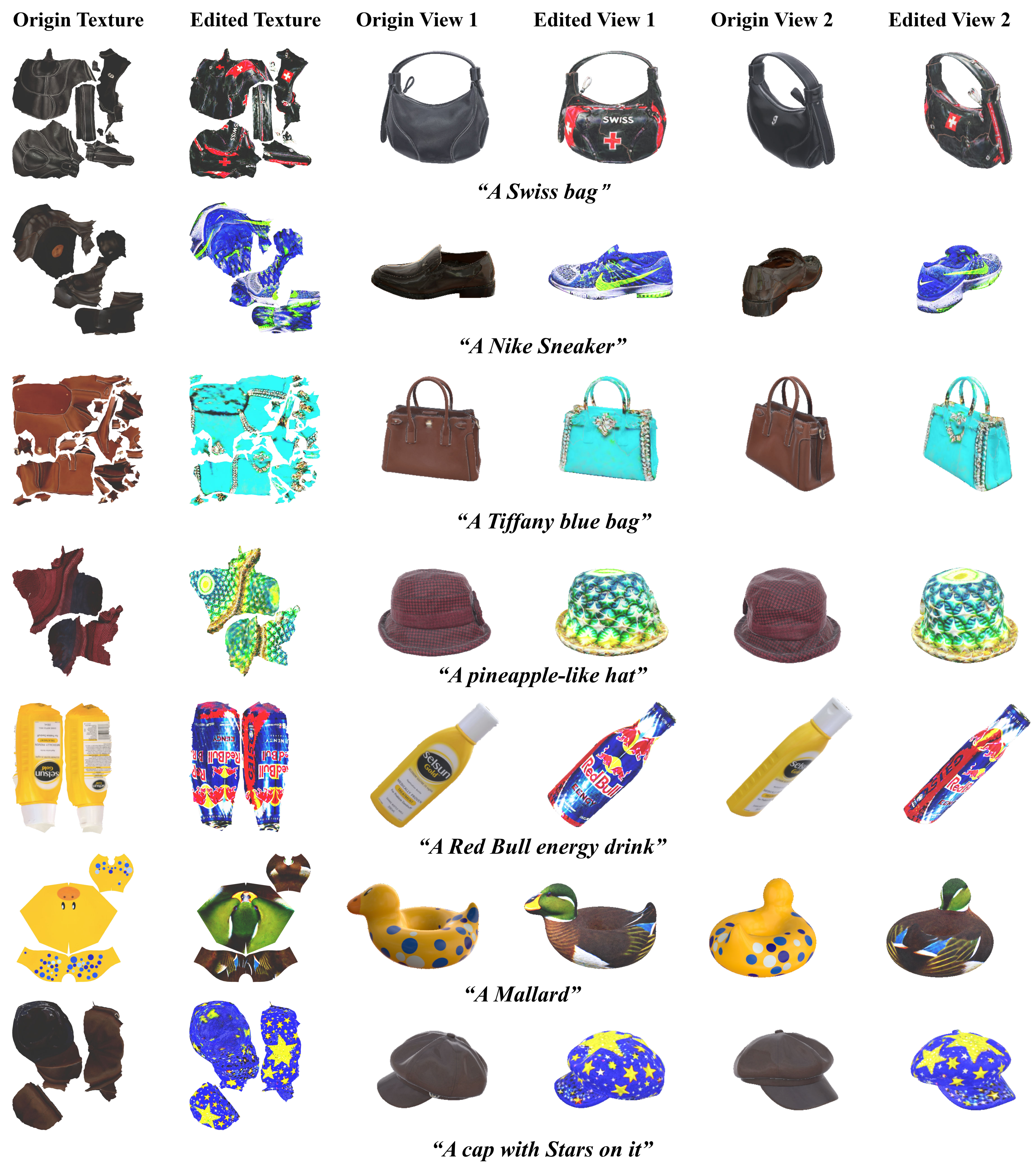}
   \caption{\textbf{Additional results} on real-world dataset. We show both textures and multi-view rendered images before and after editing.}
   \label{fig:supp_real}
\end{figure*}

\begin{figure*}[t]
  \centering
  \includegraphics[width=0.87\linewidth]{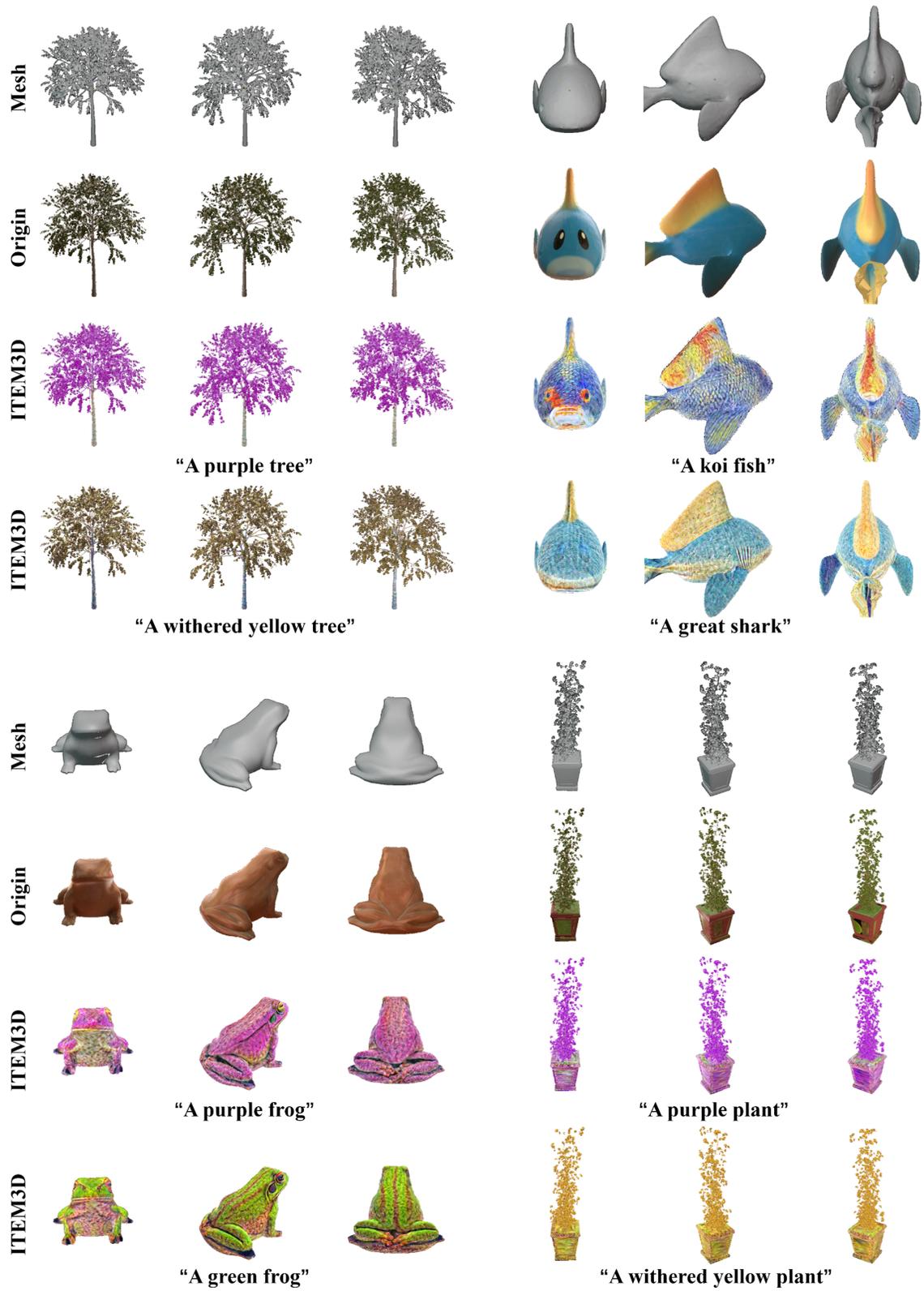}
   \caption{\textbf{Additional results} on synthetic dataset. The results of both mesh and rendered images are presented. We show the multi-view rendered images of two text prompts for each 3D object.}
   \label{fig:supp}
\end{figure*}

To begin with, we present additional qualitative results to showcase generalized editing capabilities of our ITEM3D. These results cover both real-world data and synthetic data.
In~\cref{fig:supp_real}, we conduct extensive experiments on real-world dataset, displaying multi-view rendered images and textures before and after editing. These results illustrate the generalization of our method for real-world applications.
The 3D models of these objects are also shown in our supplementary video.
In~\cref{fig:supp}, we also include additional results of our ITEM3D on synthetic dataset, consisting of both mesh and rendered images.

\section{Compare with Baselines}
\begin{figure}[t]
  \centering
  \includegraphics[width=\linewidth]{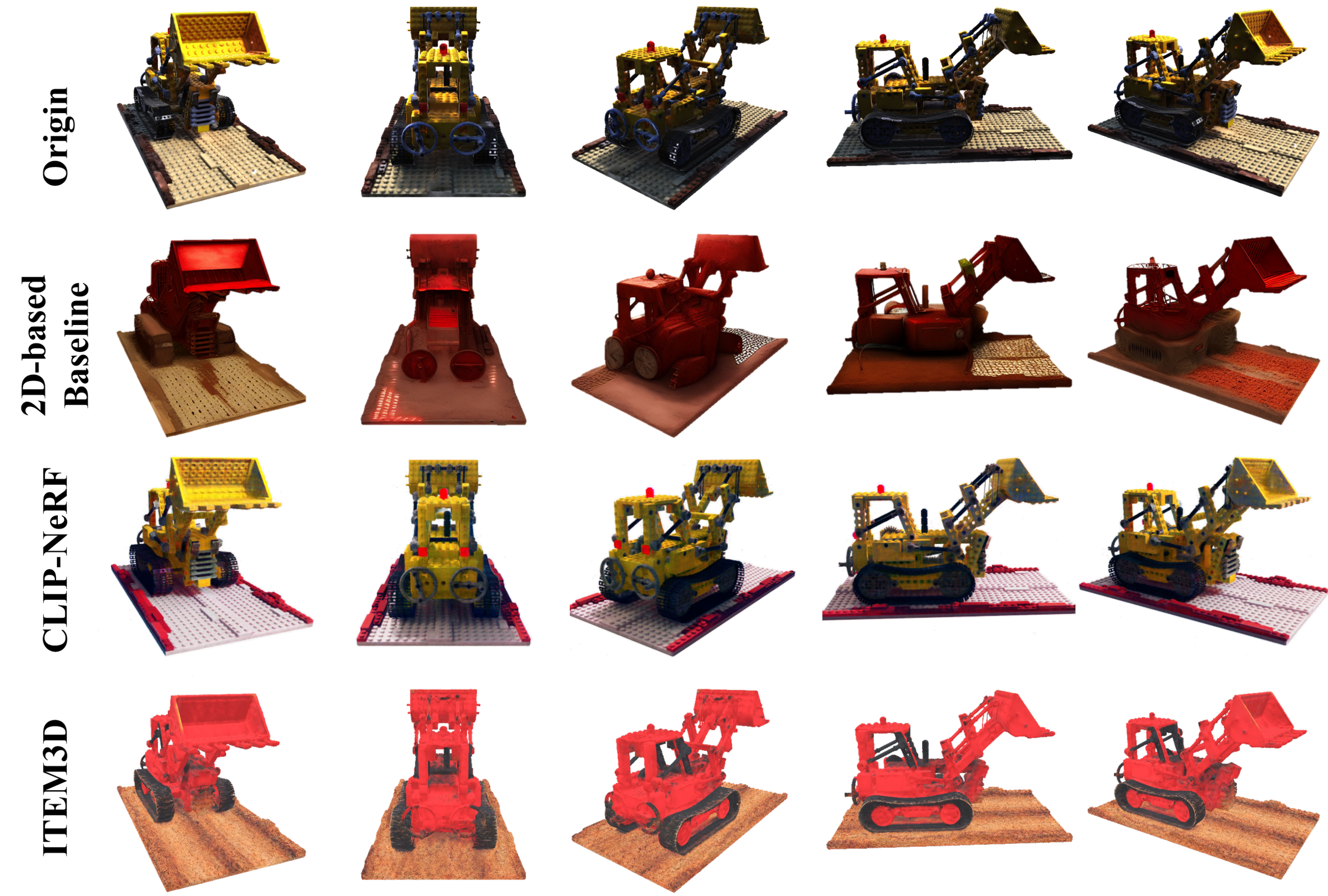}
   \caption{\textbf{Comparison with 2D baseline~\cite{zhang2023adding} and CLIP-NeRF~\cite{wang2022clip}.} We compare the performances of three methods under the prompt of \textit{``A real red excavator''}. The 2D baseline, ControlNet, successfully edits the input views into realistic red excavator, however, the multi-view results lack 3D consistency. CLIP-NeRF maintains 3D consistency, while failing to edit the color of excavator. Our ITEM3D can both edit the texture according to the prompt and keep the inherent 3D consistency.
   The collection of five-view images currently exhibits distinct patterns and slight variations in the chair's color.
  }
\label{fig:2d_baseline}
\end{figure}

To show the superiority of 3D consistency in 3D representations and knowledge prior in diffusion models, we also compare our approach with two baselines: CLIP-NeRF and a 2D baseline called ControlNet. We present the results of this comparison in~\cref{fig:2d_baseline}.


\section{Failure Case}
\begin{figure}[t]
  \centering
  \includegraphics[width=\linewidth]{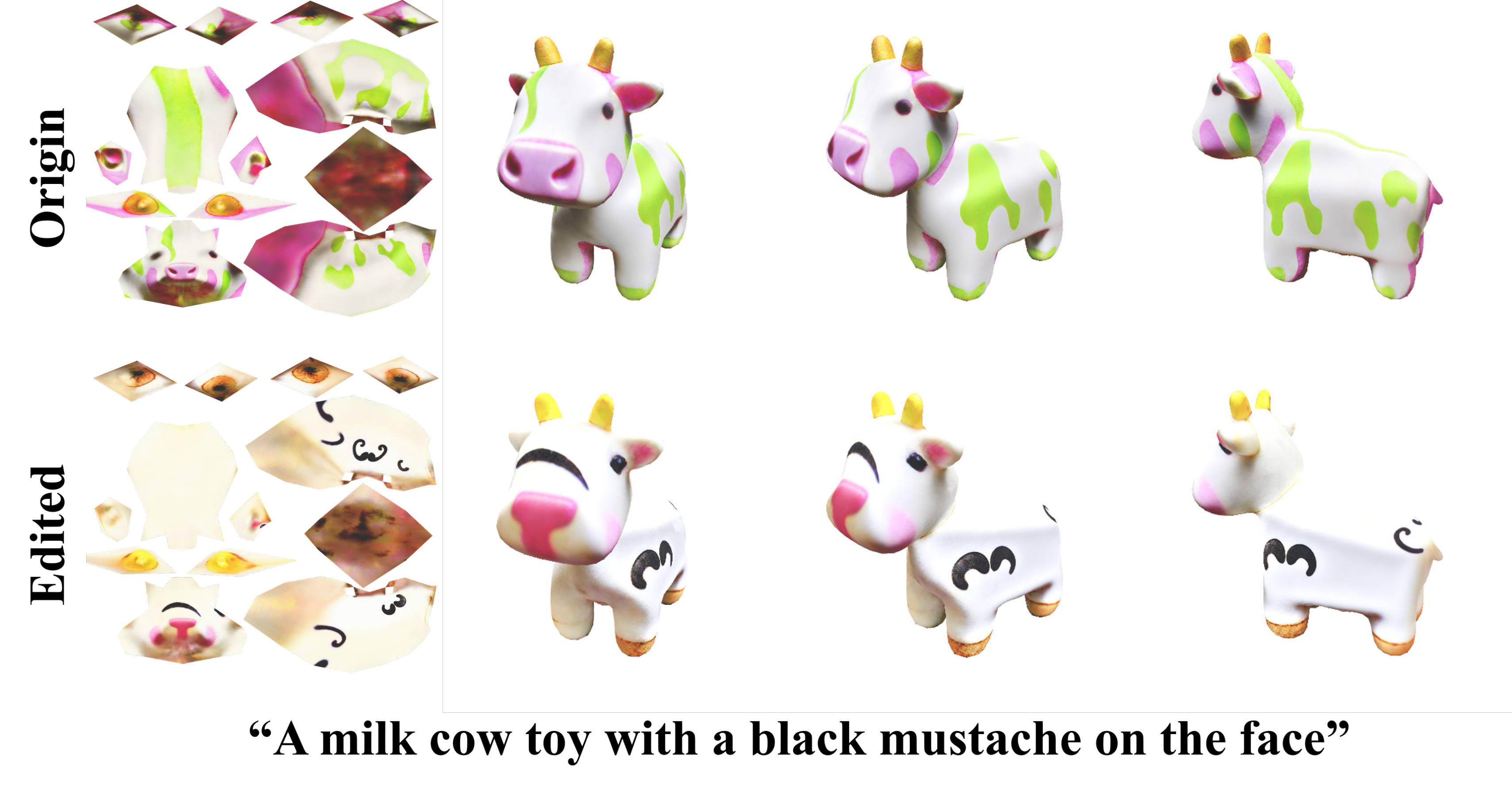}
   \caption{\textbf{Failure case.} Our method encounters challenges when dealing with complex and precise editing prompts. Our approach inadvertently alters the mustache located on the cow's body instead of its face.}
   \label{fig:complex}
\end{figure}

During the experiments, we have observed that our method fails to precisely editing the correct part of the objects for complex editing. As shown in~\cref{fig:complex}, we attempted to add a mustache to the cow. While it did generate a black mustache, it inadvertently altered the mustache located on the cow's body instead of its face.

\section{Broader Impacts}
Our ITEM3D allows for delicate editing of material properties of 3D objects while maintaining their geometric structure and UV map.
This approach has broad applications in the gaming and e-shopping industries, where object materials can be edited through textual descriptions to create various 3D objects. 
Our method can directly edit the object materials through textual descriptions, seamlessly incorporated into the industrial modeling pipeline.
However, it is crucial to know that our method carries potential risks of deception.
Our work may be used in product forgery which could become a threat to personal credit.
We do not allow any application of our work to such malicious acts.

\end{document}